%% file: neurips_2023.tex
\documentclass{article}

    \usepackage[preprint]{neurips_2023}

\usepackage[utf8]{inputenc} 
\usepackage[T1]{fontenc}    
\usepackage{hyperref}       
\usepackage{url}            
\usepackage{booktabs}       
\usepackage{amsfonts}       
\usepackage{nicefrac}       
\usepackage{microtype}      
\usepackage{xcolor}         
\usepackage{amsmath,amsfonts}
\usepackage{algorithmic}
\usepackage{algorithm}
\usepackage{array}
\usepackage[caption=false,font=normalsize,labelfont=sf,textfont=sf]{subfig}
\usepackage{textcomp}
\usepackage{stfloats}
\usepackage{url}
\usepackage{verbatim}
\usepackage{graphicx}
\usepackage{cite}
\usepackage{multirow}
\usepackage{bbding}
\usepackage{utfsym}

\usepackage{makecell}
\usepackage{soul}
\usepackage{accents}
\usepackage{enumitem}
\usepackage[nameinlink,capitalise]{cleveref}
\newcolumntype{C}[1]{>{\centering\arraybackslash}p{#1}}
\newcolumntype{L}[1]{>{\arraybackslash}p{#1}}

\title{AudioGPT: Understanding and Generating Speech, Music, Sound, and Talking Head}

\author{%
  Rongjie Huang$^{1}$\thanks{~Equal contributions}, Mingze Li$^{1}$\footnotemark[1], Dongchao Yang$^{2}$\footnotemark[1], Jiatong Shi$^{3}$\footnotemark[1], Xuankai Chang$^3$ \\ \textbf{Zhenhui Ye$^1$, Yuning Wu$^4$, Zhiqing Hong$^1$, Jiawei Huang$^1$, Jinglin Liu$^1$, Yi Ren$^1$}, \\ \textbf{Zhou Zhao$^1$, Shinji Watanabe$^3$}\\ \\ 
  Zhejiang University$^1$, Peking University$^2$, Carnegie Mellon University$^3$, Remin University of China$^4$ \\ \\ 
  \texttt{\{rongjiehuang, limingze, zhaozhou\}@zju.edu.cn}, \texttt{\{dongchao98\}@stu.pku.edu.cn}, \\ \texttt{\{jiatongs, xuankaic, dongsli\}@cs.cmu.edu}, \\ \texttt{\{yuningwu\}@ruc.edu.cn}, \texttt{\{shinjiw\}@ieee.org}
  \\ \\
\url{https://github.com/AIGC-Audio/AudioGPT} 
\vspace{-0.4cm}
  }

\begin{document}

\maketitle

\begin{abstract}
Large language models (LLMs) have exhibited remarkable capabilities across a variety of domains and tasks, challenging our understanding of learning and cognition. Despite the recent success, current LLMs are not capable of processing complex audio information or conducting spoken conversations (like Siri or Alexa). In this work, we propose a multi-modal AI system named AudioGPT, which complements LLMs (i.e., ChatGPT) with 1) foundation models to process complex audio information and solve numerous understanding and generation tasks; and 2) the input/output interface (ASR, TTS) to support spoken dialogue. With an increasing demand to evaluate multi-modal LLMs of human intention understanding and cooperation with foundation models, we outline the principles and processes and test AudioGPT in terms of consistency, capability, and robustness. Experimental results demonstrate the capabilities of AudioGPT in solving AI tasks with speech, music, sound, and talking head understanding and generation in multi-round dialogues, which empower humans to create rich and diverse audio content with unprecedented ease. 

\end{abstract}

\input{Sections/1_intro.tex}

\input{Sections/2_related.tex}
\input{Sections/3_framework.tex}
\input{Sections/3add_evaluate.tex}
\input{Sections/4_exp.tex}

\input{Sections/5_limitation.tex}

\input{Sections/6_conclusion.tex}

\bibliographystyle{neurips_2023}
\bibliography{neurips_2023}

\end{document}

%% file: Sections/1_intro.tex
\section{Introduction}
\label{sec: intro}

Nowadays, Large language models (LLMs)~\citep{devlin2018bert,raffel2020exploring,brown2020language,ouyang2022training,zhang2022opt} are posing a significant impact on the AI community, and the advent of ChatGPT and GPT-4 leads to the advancement of natural language processing. Based on the massive corpora of web-text data and powerful architecture, LLMs are empowered to read, write, and communicate like humans.

Despite the successful applications in text processing and generation, replicating this success for audio modality (speech~\citep{ren2020fastspeech,huang2022fastdiff}, music~\citep{huang2021multi,liu2022diffsinger}, sound~\citep{yang2022diffsound,huang2023make}, and talking head~\citep{wu2021imitating,ye2023geneface}) is limited, while it is highly beneficial since: 1) In real-world scenarios, humans communicate using spoken language across daily conversations, and utilize spoken assistant (e.g., Siri or Alexa) to boost life convenience; 2) As an inherent part of intelligence, processing audio modality information is a necessity to achieve artificial general intelligence. Understanding and generating speech, music, sound, and talking head could be the critical step for LLMs toward more advanced AI systems.


Despite the benefits of audio modality, training LLMs that support audio processing is still challenging due to the following issues: 1) Data: Obtaining human-labeled speech data is an expensive and time-consuming task, and there are only a few resources available that provide real-world spoken dialogues. Furthermore, the amount of data is limited compared to the vast corpora of web-text data, and multi-lingual conversational speech data is even scarcer; and 2) Computational resources: Training multi-modal LLMs from scratch is computationally intensive and time-consuming. Given that there are already existing audio foundation models that can understand and generate speech, music, sound, and talking head, it would be wasteful to start training from scratch.

\begin{figure}[!htp]
    \centering
    \caption{A high-level overview of AudioGPT. AudioGPT can be divided into four stages, including modality transformation, task analysis, model assignment, and response generation. It equips ChatGPT with audio foundation models to handle complex audio tasks and is connected with a modality transformation interface to enable spoken dialogue. We design principles to evaluate multi-modal LLMs in terms of consistency, capability, and robustness.}
    \includegraphics[width=\textwidth]{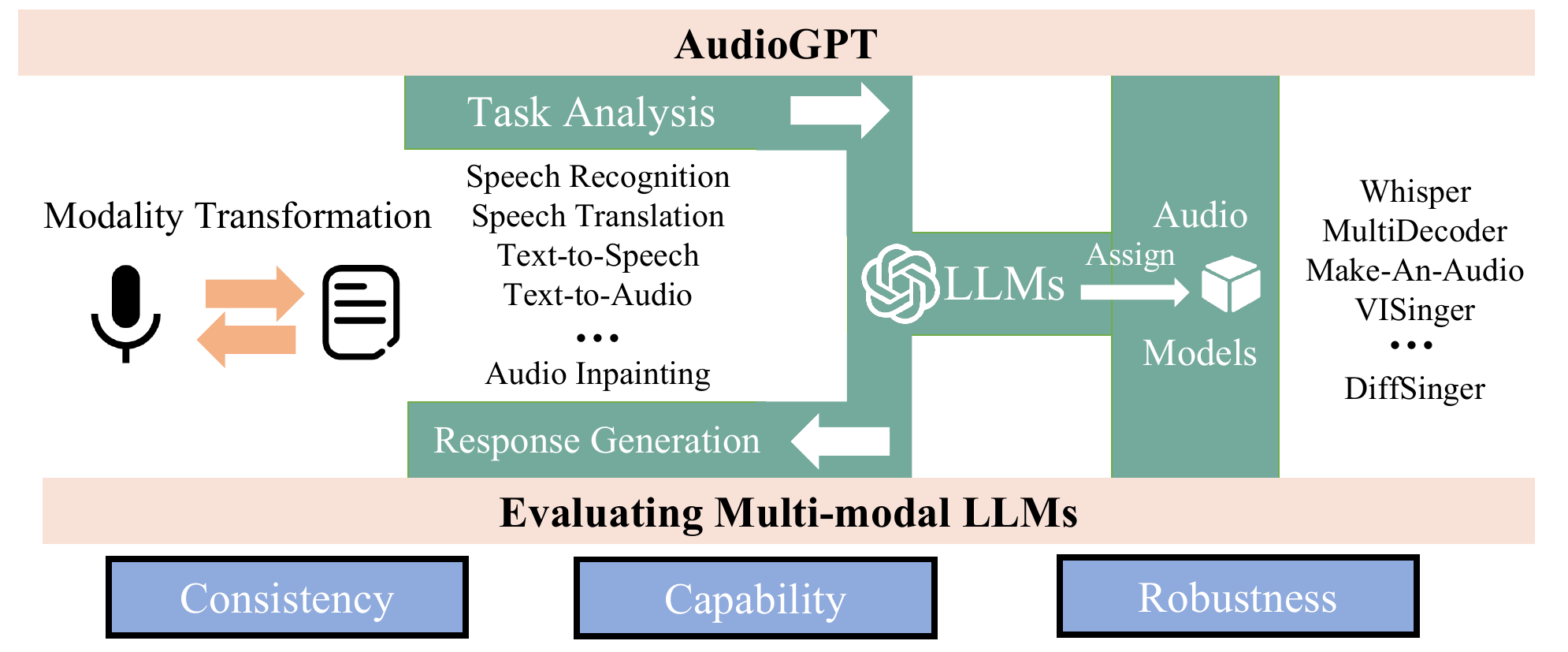}
    \label{fig:diagram}
\end{figure}

In this work, we introduce ``AudioGPT", a system designed to excel in understanding and generating audio modality in spoken dialogues. Specifically, 1) Instead of training multi-modal LLMs from scratch, we leverage a variety of audio foundation models to process complex audio information, where LLMs (i.e., ChatGPT) are regarded as the general-purpose interface~\citep{wu2023visual,shen2023hugginggpt} which empowers AudioGPT to solve numerous audio understanding and generation tasks; 2) Instead of training a spoken language model, we connect LLMs with input/output interface (ASR, TTS) for speech conversations; As illustrated in Figure~\ref{fig:diagram}, the whole process of AudioGPT can be divided into four stages:

\begin{itemize}
    \item Modality Transformation. Using input/output interface for modality transformation between speech and text, bridging the gap between the spoken language LLMs and ChatGPT.
    \item Task Analysis. Utilizing the dialogue engine and prompt manager to help ChatGPT understands the intention of a user to process audio information. 
    \item Model Assignment. Receiving the structured arguments for prosody, timbre, and language control, ChatGPT assigns the audio foundation models for understanding and generation.
    \item Response Generation. Generating and returning a final response to users after the execution of audio foundation models.
\end{itemize}

As a blossoming research topics~\citep{wu2023visual,shen2023hugginggpt,huang2023language}, there is an increasing demand for evaluating the performance of multi-modal LLMs in understanding human intention and organizing the cooperation of multiple foundation models. In this work, we outline the design principles and process of evaluating AudioGPT in terms of consistency, capability, and robustness. Experimental results demonstrate the capabilities of AudioGPT for processing complex audio information in multi-round dialogue, covering a series of AI tasks including generating and understanding speech, music, sound, and talking head.


Key contributions of the paper include:
\begin{itemize}
    \item We propose AudioGPT, which equips ChatGPT with audio foundation models to handle complex audio tasks. As a general-purpose interface, ChatGPT is connected with a modality transformation interface to enable spoken dialogue. 
    \item We outline the design principles and process of evaluating multi-modal LLMs, and test AudioGPT in terms of consistency, capability, and robustness.  
    \item Demonstrations present the efficiency of AudioGPT in audio understanding and generation with multiple rounds of dialogue, which empowers humans to create rich and diverse audio content with unprecedented ease.
\end{itemize}

%% file: Sections/2_related.tex
\section{Related Works}
\label{sec: related}

\subsection{Large Language Models}
The research areas of AI are being revolutionized by the rapid progress of Large Language Models (LLMs)~\citep{brown2020language,ouyang2022training,zhang2022opt}, where they can serve as a general-purpose language task solver, and the research paradigm has been shifting towards the use of LLMs. They have long been considered a core problem in natural language processing and demonstrated remarkable abilities for tasks such as machine translation~\citep{gulcehre2017integrating,baziotis2020language}, open-ended dialogue modeling~\citep{hosseini2020simple,thoppilan2022lamda}, and even code completion~\citep{svyatkovskiy2019pythia,liu2020multi}. 

Among them, ~\citet{kaplan2020scaling} studied the impact of scaling on the performance of deep learning models, showing the existence of power laws between the model and dataset sizes and the performance of the system. Language models (LMs) at scale, such as GPT-3~\citep{brown2020language} have demonstrated remarkable performance in few-shot learning. FLAN~\citep{wei2021finetuned} is proposed to improve the zero-shot performance of large language models, which would expand their reach to a broader audience. LLaMA~\citep{touvron2023llama} shows that it is possible to achieve state-of-the-art performance by training exclusively on publicly available data, without resorting to proprietary datasets. The advent of ChatGPT and GPT-4 leads to rethinking the possibilities of artificial general intelligence (AGI).

\subsection{Spoken Generative Language Models}

Self-supervised learning (SSL) has emerged as a popular solution to many speech processing problems with a massive amount of unlabeled speech data. HuBERT~\citep{hsu2021hubert} is trained with a masked prediction with masked continuous audio signals. Inspired by vector quantization (VQ) techniques, SoundStream~\citep{zeghidour2021soundstream} and Encodec~\citep{defossez2022high} present the hierarchical architecture for high-level representations that carry semantic information.

Most of these models build discrete units in a compact and discrete space, which could be modeled with an autoregressive Transformer whose predictions are then mapped back to the original signal space. ~\citet{hayashi2020discretalk} leverage discrete VQ-VAE representations to build speech synthesis models via autoregressive machine translation. ``textless NLP''~\citep{kharitonov2022textless,huang2022transpeech} is proposed to model language directly without any transcription by training autoregressive generative models of low-bitrate audio tokens. AudioLM~\citep{borsos2022audiolm} and MusicLM~\citep{agostinelli2023musiclm} follow a similar way to address the trade-off between coherence and high-quality synthesis, where they cast audio synthesis as a language modeling task and leverage a hierarchy of coarse-to-fine audio discrete units in a discrete representation space.

Recently, ~\citet{nguyen2023generative} leverage the success of discrete representation and introduce the first end-to-end generative spoken dialogue language model. However, due to the data and computational resource scarcity mentioned above, it would be challenging to train spoken generative language models from scratch that enables the processing of complex audio information. Differently, we regard LLMs (i.e., ChatGPT) as the general-purpose interface and leverage various audio foundational models to solve audio understanding and generation tasks, where AudioGPT is further connected with modality transformation to support speech conversations.

%% file: Sections/3_framework.tex
\section{AudioGPT}

\subsection{System Formulation}
\label{ssec: formulation}

As briefly discussed in Sec.~\ref{sec: intro}, AudioGPT is a prompt-based system, defined as 
\begin{equation}
    \mathrm{AudioGPT} = (\mathcal{T}, \mathcal{L}, \mathcal{M}, \mathcal{H}, \{\mathcal{P}_i\}_{i=1}^{P}),
\end{equation}
where $\mathcal{T}$ is a modality transformer, $\mathcal{L}$ is a dialogue engine (i.e., large language model, LLM), $\mathcal{M}$ is a prompt manager, $\mathcal{H}$ is a task handler, and $\{\mathcal{P}_i\}_{i=1}^{P}$ is a set of $P$ audio foundation models. Let a context with $(n-1)$-rounds interactions to be defined as $C = \{(q_1, r_1), (q_2, r_2), ..., (q_{n-1}, r_{n-1}))\}$, where $q_i$ is the query of $i^{\text{th}}$ round and $r_i$ is the response of $i^{\text{th}}$ round. Denoted a new query $q_n$, the execution of the AudioGPT is to generate the response $r_n$ as formulated in:
\begin{equation}
    r_n = \mathrm{AudioGPT}(q_n, C)
\end{equation}


During inference, AudioGPT can be decomposed into four major steps: 1) \textbf{Modality transformation}: transfer various input modalities within $q_n$ into a query $q_n'$ with a consistent modality; 2) \textbf{Task analysis}: utilize the dialogue engine $\mathcal{L}$ and the prompt manager $\mathcal{M}$ to parse $(q_n', C)$ into structure arguments $a_n$ for the task handler $\mathcal{H}$; 3) \textbf{Model assignment}: the task handler $\mathcal{H}$ consumes structured arguments $a_n$ and send the arguments to its corresponding audio task processor $\mathcal{P}_s$, where $s$ is the selected task index, and 4) \textbf{response generation}: after execution of $\mathcal{P}_s(a_n)$, the final response $r_n$ is generated through $\mathcal{L}$ by combining information from $(q'_n, C, \mathcal{P}_s(a_n))$.

\subsection{Modality Transformation}

As discussed in Sec.~\ref{ssec: formulation}, the first stage aims to transform the query $q_n$ into a new query $q'_n$ in a consistent format. The user input query $q_n$ includes two parts: a query description $q^{(d)}_n$ and a set of query-related resources of size $k$, $\{q^{(s_1)}_n, q^{(s_2)}_n, ..., q^{(s_k)}_n\}$. In AudioGPT, the query description $q^{(d)}_n$ can be either in textual or audio (i.e., speech) format. And the modality transformer $\mathcal{T}$ first checks the modality of query description $q^{(d)}_n$. If the query description $q^{(d)}_n$ is in audio, $\mathcal{T}$ is then responsible for converting $q^{(d)}_n$ in audio to textual modality as:
\begin{equation}
    q'_n = (q'^{(d)}_n, \{q^{(s_1)}_n, ..., q^{(s_k)}_n\}) = 
    \begin{cases}
        (q^{(d)}_n, \{q^{(s_1)}_n, ..., q^{(s_k)}_n\}) & \text{if } q^{(d)}_n \text{ is text,} \\
        (\mathcal{T}(q^{(d)}_n), \{q^{(s_1)}_n, ..., q^{(s_k)}_n\}) & \text{if } q^{(d)}_n \text{ is audio.} \\
    \end{cases}
\end{equation}

\subsection{Task Analysis}

As introduced in Sec.~\ref{ssec: formulation}, the task analysis step focuses on extracting structured argument $a_n$ from $(q'_n, C)$. Specifically, the context $C$
is fed into the dialogue engine $\mathcal{L}$ ahead of the argument extraction. Based on the types of query resources $\{q^{(s_1)}_n, q^{(s_2)}_n, ..., q^{(s_k)}_n\}$ from $q'_n$, the task handler $\mathcal{H}$ first classifies the query into different task families, which is classified through I/O modalities. Then, given the task family selected, the query description $q'^{(d)}_n$ is passed into the prompt manager $\mathcal{M}$ to generate argument $a_n$, including the selected audio foundation model $\mathcal{P}_p$ and its corresponding task-related arguments $h_{\mathcal{P}_p}$, where $p$ is the index of  the selected audio model from the audio model set $\{\mathcal{P}_i\}_{i=1}^{P}$. 

\begin{equation}
    (\mathcal{P}_p, h_{\mathcal{P}_p}) = \mathcal{L}(\mathcal{M}(\mathcal{H}(q'_n), q'^{(d)}_n), C),
\end{equation}
where $\mathcal{H}(q'_n)$ is the task family selected by the task handler $\mathcal{H}$. Noted that, for an audio/image-input task family, $h_{\mathcal{P}_p}$ may also contain the necessary resources (e.g., audio or images) from the previous context $C$.

\begin{table}[]
    \centering
        \caption{Supported Tasks in AudioGPT}
\resizebox {\linewidth} {!} {
    \begin{tabular}{l|cc|cl}
        \toprule
        Task & Input & Output & Domain & Model \\
        \midrule
        Speech Recognition & Audio & Text & Speech & Whisper \citep{radford2022robust} \\
        Speech Translation & Audio & Text & Speech & MultiDecoder \citep{dalmia2021searchable} \\
        \midrule
        Style Transfer & Audio & Audio & Speech & GenerSpeech \citep{huang2022generspeech} \\
        Speech Enhancement & Audio & Audio & Speech & ConvTasNet \citep{luo2019conv} \\
        Speech Separation & Audio & Audio & Speech & TF-GridNet \citep{wang2022tf} \\
        Mono-to-Binaural & Audio & Audio & Speech & NeuralWarp \citep{grabocka2018neuralwarp} \\
        Audio Inpainting & Audio & Audio & Sound & Make-An-Audio \citep{huang2023make} \\
        Sound Extraction & Audio & Audio & Sound & LASSNet \citep{liu2022separate} \\ \midrule
        Sound Detection & Audio & Event & Sound & Pyramid Transformer \citep{xin2022audio} \\
        \midrule
        Talking Head Synthesis & Audio & Video & Talking Head & GeneFace \citep{ye2023geneface} \\
        \midrule
        Text-to-Speech & Text & Audio & Speech & FastSpeech 2 \citep{ren2020fastspeech} \\
        Text-to-Audio & Text & Audio & Sound & Make-An-Audio \citep{huang2023make}\\  
        \midrule
        Audio-to-Text & Audio &Text  &Sound  & MAAC \citep{ye2021improving} \\
        \midrule
        Image-to-Audio & Image & Audio & Sound & Make-An-Audio \citep{huang2023make} \\
        \midrule
        \multirow{2}{*}{Singing Synthesis} & \multirow{2}{*}{Musical Score}  & \multirow{2}{*}{Audio}  & \multirow{2}{*}{Music} & DiffSinger \citep{liu2022diffsinger} \\
        &  &  &  & VISinger \citep{zhang2022visinger} \\
        \bottomrule
    \end{tabular}}

    \label{tab:my_label}
\end{table}

As aforementioned, the task family is determined through the task handler $\mathcal{H}$ by considering the I/O modality. To be specific, the families are:

\begin{itemize}
        \item Audio-to-Text \begin{itemize}
            \item Speech Recognition: Transcribe human speech
            \item Speech Translation: Translate human speech into another language
            \item Audio Caption: Describe audio in text
        \end{itemize}
        \item Audio-to-Audio \begin{itemize}
            \item Style Transfer: Generate human speech with styles derived from a reference
            \item Speech Enhancement: Improve the speech quality by reducing background noise
            \item Speech Separation: Separate mix-speech of different speakers
            \item Mono-to-Binaural: Generate binaural audio given mono one
            \item Audio Impainting: Inpaint audio given user input mask
        \end{itemize}
        \item Audio-to-Event \begin{itemize}
            \item Sound Extraction: Selectly extract a part of audio based on description
            \item Sound Detection: Predict the event timelines in audio
        \end{itemize}
        \item Audio-to-Video \begin{itemize}
            \item Talking Head Synthesis: Generate a talking human portrait video given input audio
        \end{itemize}
        \item Text-to-Audio \begin{itemize}
            \item Text-to-Speech: Generate human speech given user input text
            \item Text-to-Audio: Generate general audio given user description
        \end{itemize}
        \item Image-to-Audio \begin{itemize}
            \item Image-to-Audio: Generate audio from image
        \end{itemize}
        \item Score-to-Audio \begin{itemize}
            \item Singing Synthesis: Generate singing voice given input text, note and duration Sequence
        \end{itemize}
\end{itemize}

\subsection{Model Assignment}
Given the selected model $\mathcal{P}_p$ and its corresponding arguments $h_{\mathcal{P}_p}$, this step assigns the related resources to the model and executes the model $\mathcal{P}_p$ to get the task output $o_{\mathcal{P}_p}$:
\begin{equation}
    o_{\mathcal{P}_p} = \mathcal{P}_p(\{q^{(s_1)}_n, q^{(s_2)}_n, ..., q^{(s_k)}_n\}, h_{\mathcal{P}_p}).
\end{equation}

To keep the efficiency of AudioGPT, we conduct the audio model initialization during either environmental setups or server initialization.

\subsection{Response Generation}
The response generation is highly related to the select task $\mathcal{P}_p$ and its output $ o_{\mathcal{P}_p}$. Specifically, for audio generation tasks, AudioGPT shows both the waveform in an image and the corresponding audio file for downloading/playing; for tasks that generate text, the model directly returns the transcribed text; for the video generation task, the output video and some related image frames are shown; for classification tasks, a posteriorgram of categories is shown over the time span.

%% file: Sections/3add_evaluate.tex
\section{Evaluating Multi-Modal LLMs}

\subsection{Overview}

The rapid development of multi-modal LLMs~\citep{wu2023visual,shen2023hugginggpt,huang2023language} has significantly increased the research demand for evaluating its performance and behavior in understanding human intention, performing complex reasoning, and organizing the cooperation of multiple audio foundation models. 

In this section, we outline the design principles and process of evaluating multi-modal LLMs (i.e., AudioGPT). Specifically, we evaluate the LLMs in the following three aspects: 1) Consistency, which measures whether the LLMs properly understand the intention of a user, and assigns the audio foundation models closely aligned with human cognition and problem-solving; 2) Capabilitity, which measures the performance of audio foundation models in handling complex audio tasks, understanding and generating speech, music, sound, and talking head in a zero-shot fashion; and 3) Robustness, which measures the ability of LLMs deals with special cases.

\subsection{Consistency}
\label{ssec: consistency}

\begin{figure}[!htp]
    \centering
    \caption{A high-level overview of consistency evaluation.}
    \includegraphics[width=\textwidth]{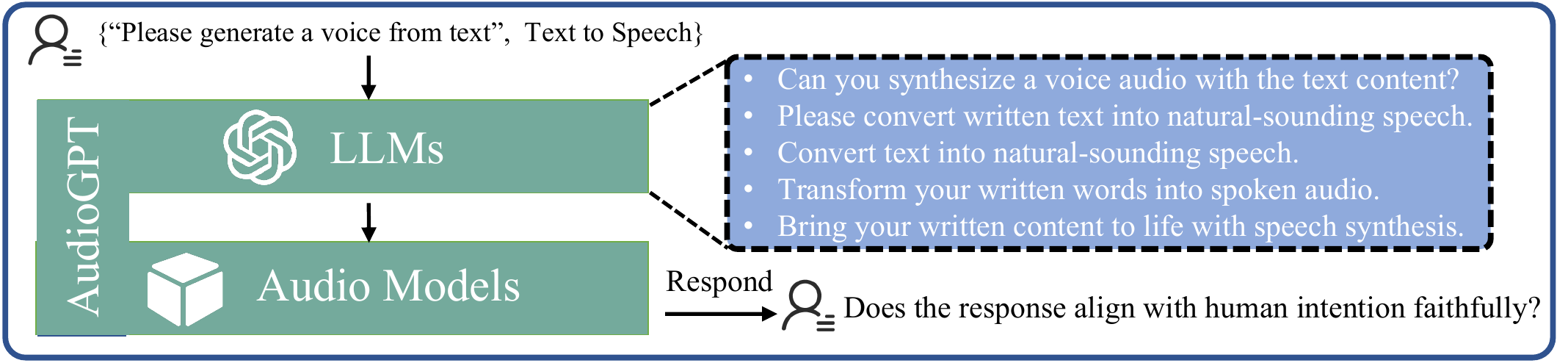}
    \label{fig:evaluation}
\end{figure} 

\begin{table}[ht]
    \centering
   \small
       \caption{Ratings that have been used in the evaluation of query-answer consistency.}
       \vspace{2mm}
     \begin{tabular}{ccc}
     \toprule
     Rating & Consistency & Definition                           \\
     \midrule
     20     & Completely inconsistent     &  Very annoying and objectionable inconsistency. \\
     40     & Mostly inconsistent      &  Annoying but not objectionable inconsistency. \\
     60     & somewhat consistent       &  Perceptible and slightly annoying inconsistency\\
     80     & Mostly consistent          & Just perceptible but not annoying inconsistency. \\
     100    & Completely consistent      & Imperceptible inconsistency\\
     \bottomrule
     \end{tabular}
     \label{matrix:naturalness}
     \end{table}

In the consistency evaluation for the zero-shot setting, models are directly evaluated on the questions without being provided any prior examples of the specific tasks, which evaluate whether multi-modal LLMS could reason and solve problems without explicit training.

More specifically, as shown in Figure~\ref{fig:evaluation}, the consistency evaluation is carried out in three steps for each task in the benchmark. In the first step, we request human annotators to provide prompts for each task in a format of \{prompts, task\_name\}. This allows us to evaluate the model’s ability to comprehend complex tasks and identify the essential prompts needed for successful task assignments. In the second step, we leverage the outperformed language generation capacity of LLMs to produce descriptions with the same semantic meanings while having different expressions, enabling a comprehensive evaluation of whether LLMs understands the intention of a broader amount of user. Finally, we use crowd-sourced human evaluation via Amazon Mechanical Turk, where AudioGPT is prompted with these natural language descriptions corresponding to a variety of tasks and intentions. Human raters are shown the response of multi-modal LLMs and a prompt input and asked ``Does the response closely align with human cognition and intention faithfully?". They must respond with ``completely", ``mostly", or ``somewhat" on a 20-100 Likert scale, which is documented with 95\% confidence intervals (CI).

\subsection{Capability}

As the task executors for processing complex audio information, audio foundation models have a significant impact on handling complex downstream tasks. Taking AudioGPT as an example, we report evaluation matrics and downstream datasets for understanding and generating speech, music, sound, and talking head in Table~\ref{tab:evaluate_capabilitity}. 

\begin{table}[ht]
    \centering
        \caption{Evaluating audio foundation models in AudioGPT.}
\resizebox {0.9\linewidth}{!} {
    \begin{tabular}{l|c|cc}
        \toprule
        Task & Audio Model & Dataset & Metrics  \\
        \midrule
        Speech Recognition& Whisper           & LibriTTS  &  WER, CER     \\
        Speech Translation& MultiDecoder      &   MUSTC   &  BLEU   \\
        \midrule
        Style Transfer    & GenerSpeech       & ESD       & MCD, FFE, MOS  \\
        Speech Enhancement  & ConvTasNet      &   CHiME4             &  SNR, PESQ, STOI  \\
        Speech Separation  & TF-GridNet       & WSJ0-2mix                        &  SNR, PESQ, STOI  \\
        Mono-to-Binaural  & NeuralWarp        &   BinauralDataset             &  L2 Error, PESQ, MRSTFT     \\
        Audio Inpainting & Make-An-Audio      &  AudioCaption  &  MOS  \\
        Sound Extraction & LASSNet            &   AudioCaption             & SNR, PESQ   \\
        \midrule
        Sound Detection        & Pyramid Transformer  & AudioSet    & mAP    \\
        Target Sound Detection &  TSDNet                    & URBAN-SED    &  F-score   \\
        \midrule
        Talking Head Synthesis & GeneFace    & LRS3-TED        & FID, LMD    \\
        \midrule
        Text-to-Speech & Fastspeech2        & LJSpeech       & MCD, FFE, MOS      \\
        Text-to-Audio  & Make-an-Audio       & AudioCaption   & FID, KL, CLAP, MOS     \\
        Audio-to-Text  & MAAC       & Clotho-v2   & CIDEr-D     \\
        \midrule
        Image-to-Audio & Make-An-Audio      & AudioCaption    & MOS        \\
        \midrule
        Singing Synthesis & DiffSinger, VISinger  & OpencPOP & MCD, FFE, MOS     \\
        \bottomrule
    \end{tabular}}
    \label{tab:evaluate_capabilitity}
\end{table}

\subsection{Robustness}

We evaluate the robustness of multi-modal LLMs by assessing their ability to handle special cases. These cases can be classified into the following categories:
\begin{itemize}
    \item Long chains of evaluation: Multi-modal LLMs are expected to handle long chains of evaluation while considering short and long context dependencies in multi-modal generation and reuse. A chain of tasks can be presented either as a query that requires sequential application of candidate audio models, as consecutive queries that ask for different tasks, or as a mixture of the two types.
    \item Unsupported tasks: Multi-modal LLMs should be able to provide reasonable feedback to queries that require unsupported tasks not covered by the foundation models.
    \item Error handling of multi-modal models: Multi-modal foundation models can fail due to different reasons, such as unsupported arguments or unsupported input formats. In such scenarios, multi-modal LLMs need to provide reasonable feedback to queries that explain the encountered issue and suggest potential solutions.
    \item Breaks in context: Multi-modal LLMs are expected to process queries that are not in a logical sequence. For instance, the user may submit random queries in a query sequence but continue to proceed with previous queries that have more tasks.
\end{itemize}

To evaluate the robustness, we conduct a three-step subjective user rating process, similar to the steps discussed in Sec.\ref{ssec: consistency}. In the first step, human annotators provide prompts based on the above four categories. In the second step, the prompts are fed into the LLM to formulate a complete interaction session. Finally, a different set of subjects recruited from multi-modal LLMs rate the interaction on the same 20-100 scale as described in Sec.\ref{ssec: consistency}.

%% file: Sections/4_exp.tex
\section{Experiments}

\subsection{Experimental Setup}

In our experiments, we employ the gpt-3.5-turbo of the GPT models as the large language models and guide the LLM with LangChain~\citep{Chase_LangChain_2022}. The deployment of the audio foundation models requires only a flexible NVIDIA T4 GPU on hugging face space. We use a temperature of zero to generate output using greedy search and set the maximum number of tokens for generation to 2048. The current manuscript mainly covers the system description, where the experiments are designed more for demonstration.

\begin{figure}[!htp]
    \centering
    \includegraphics[width=\textwidth]{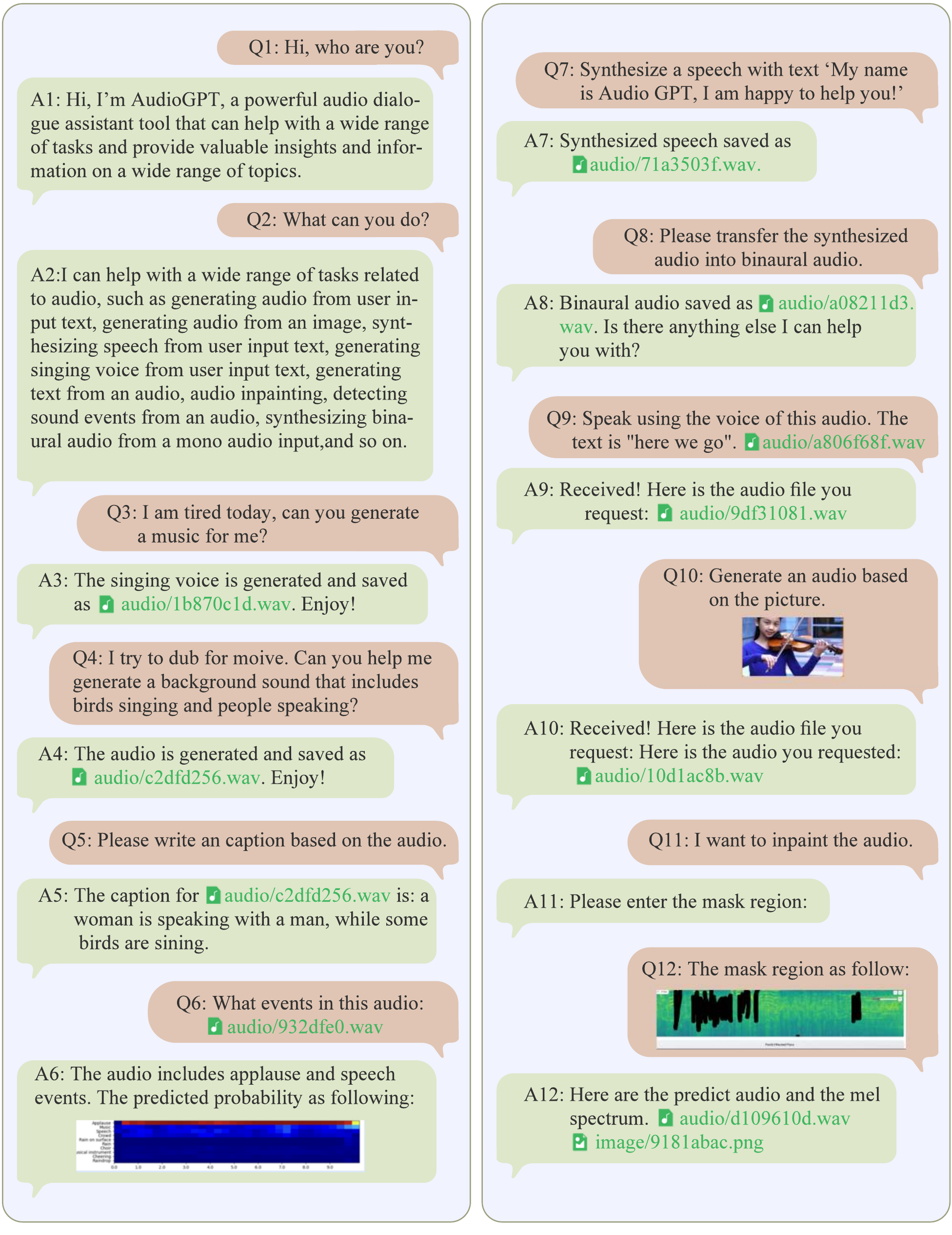}
    \caption{Qualitative analysis on multiple rounds of dialogue between humans and AudioGPT. }
    \label{exp1}
\end{figure} 
\begin{figure}[!htp]
    \centering
    \includegraphics[width=0.9\textwidth]{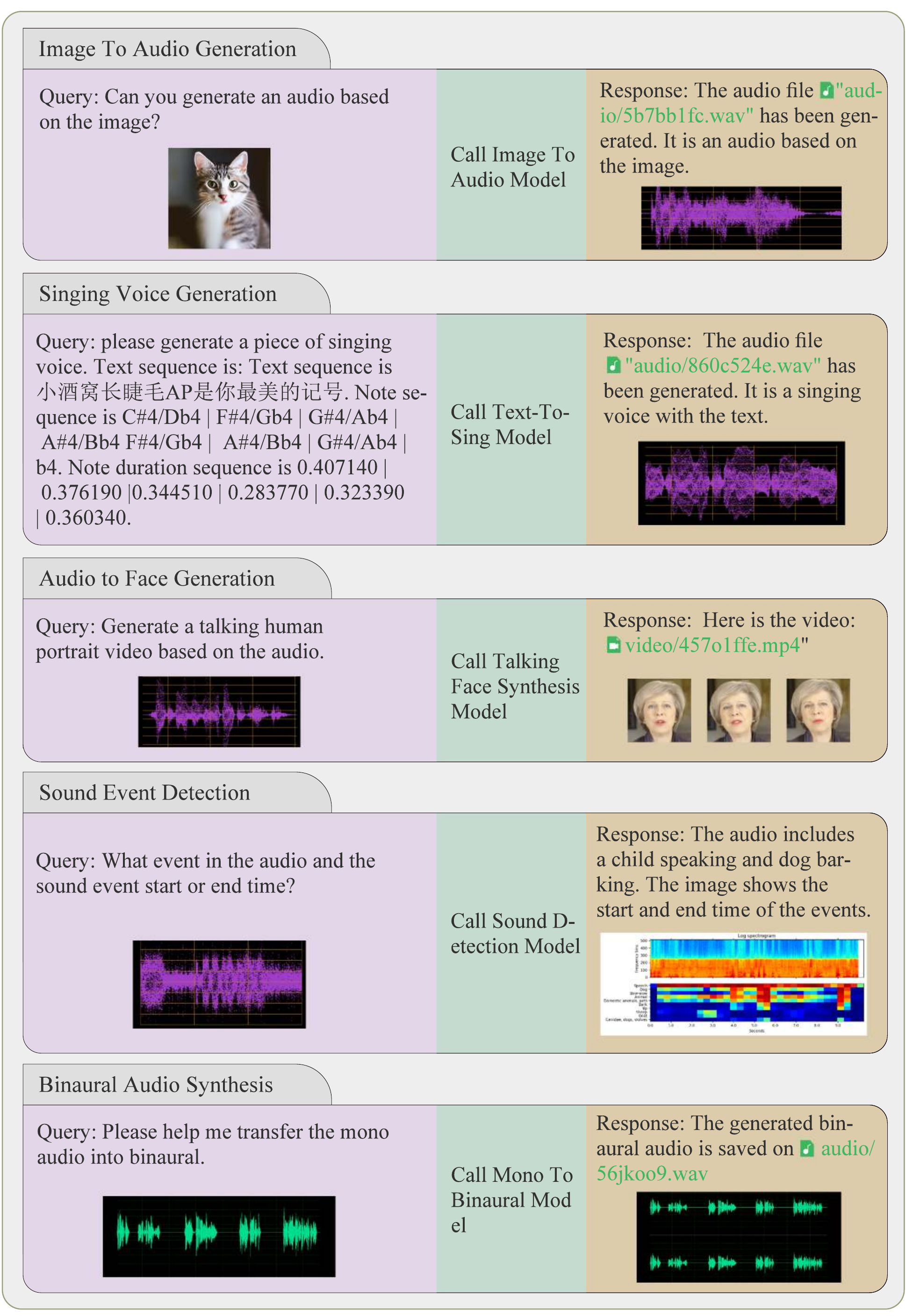}
    \caption{Qualitative analysis on simple tasks.}
    \label{exp2}
\end{figure} 
\begin{figure}[!htp]
    \centering
    \includegraphics[width=0.9\textwidth]{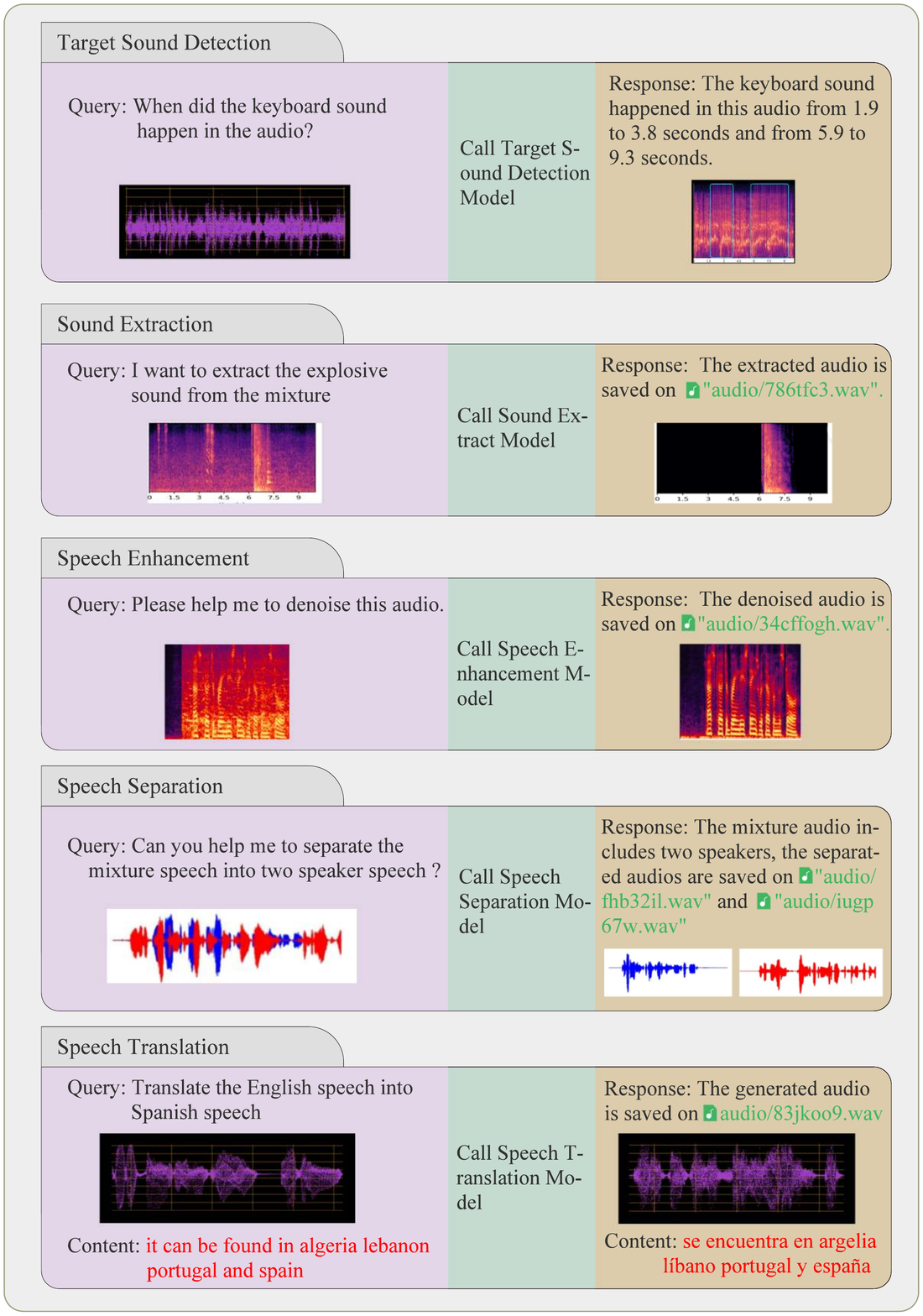}
    \caption{Qualitative analysis on simple tasks.}
    \label{exp3}
\end{figure} 

\subsection{Case Study on Multiple Rounds Dialogue}
Figure~\ref{exp1} shows a 12-rounds dialogue case of AudioGPT, which demonstrates the capabilities of AudioGPT for processing audio modality, covering a series of AI tasks in generating and understanding speech, music, sound, and talking head. The dialogue involves multiple requests to process audio information and shows that AudioGPT maintains the context of the current conversation, handles follow-up questions, and interacts with users actively. 

\subsection{Case Study on Simple Tasks}
AudioGPT equips ChatGPT with audio foundation models, where ChatGPT is regarded as the general-purpose interface to solve numerous audio understanding and generation tasks. We test AudioGPT on a wide range of audio tasks in generating and understanding speech, music, sound, and talking head, where some cases are illustrated in Figure \ref{exp2} and \ref{exp3}.

        


%% file: Sections/5_limitation.tex
\section{Limitation}
Although AudioGPT excels at solving complex audio-related AI tasks, limitations could be observed in this system as follows: 1) \textbf{Prompt Engineering}: AudioGPT uses ChatGPT to connect a large number of foundation models, and thus it requires prompt engineering to describe audio foundation models in natural language, which could be time-consuming and expertise-required; 2) \textbf{Length Limitation}: the maximum token length in ChatGPT may limit the multi-turn dialogue, which also influences the user's context instruction, and 3) \textbf{Capabliity Limitation} AudioGPT relies heavily on audio foundation models to process audio information, which is heavily influenced by the accuracy and effectiveness of these models.

%% file: Sections/6_conclusion.tex
\section{Conclusion}
In this work, we presented AudioGPT, which connected ChatGPT with 1) audio foundation models to handle challenging audio tasks, and 2) a modality transformation interface to enable spoken dialogue. By combining the advantages of ChatGPT and audio-modality solvers, AudioGPT presented strong capacities in processing audio information in the following four stages: modality transformation, task analysis, model assignment, and response generation. To assess the ability of multi-modal LLMs in human intention understanding and cooperation with foundation models, we outlined the design principles and processes, and evaluated AudioGPT in terms of consistency, capability, and robustness. Experimental results demonstrated the outperformed abilities of AudioGPT in solving AI tasks with speech, music, sound, and talking head understanding and generation in multi-round dialogues, empowering humans to create rich and diverse audio content with unprecedented ease. The current manuscript mainly covers the system description, where the experiments are designed more for demonstration.

%% file: neurips_2023.bbl
\begin{thebibliography}{45}
\providecommand{\natexlab}[1]{#1}
\providecommand{\url}[1]{\texttt{#1}}
\expandafter\ifx\csname urlstyle\endcsname\relax
  \providecommand{\doi}[1]{doi: #1}\else
  \providecommand{\doi}{doi: \begingroup \urlstyle{rm}\Url}\fi

\bibitem[Agostinelli et~al.(2023)Agostinelli, Denk, Borsos, Engel, Verzetti,
  Caillon, Huang, Jansen, Roberts, Tagliasacchi,
  et~al.]{agostinelli2023musiclm}
Agostinelli, A., Denk, T.~I., Borsos, Z., Engel, J., Verzetti, M., Caillon, A.,
  Huang, Q., Jansen, A., Roberts, A., Tagliasacchi, M., et~al.
\newblock Musiclm: Generating music from text.
\newblock \emph{arXiv preprint arXiv:2301.11325}, 2023.

\bibitem[Baziotis et~al.(2020)Baziotis, Haddow, and
  Birch]{baziotis2020language}
Baziotis, C., Haddow, B., and Birch, A.
\newblock Language model prior for low-resource neural machine translation.
\newblock \emph{arXiv preprint arXiv:2004.14928}, 2020.

\bibitem[Borsos et~al.(2022)Borsos, Marinier, Vincent, Kharitonov, Pietquin,
  Sharifi, Teboul, Grangier, Tagliasacchi, and Zeghidour]{borsos2022audiolm}
Borsos, Z., Marinier, R., Vincent, D., Kharitonov, E., Pietquin, O., Sharifi,
  M., Teboul, O., Grangier, D., Tagliasacchi, M., and Zeghidour, N.
\newblock Audiolm: a language modeling approach to audio generation.
\newblock \emph{arXiv preprint arXiv:2209.03143}, 2022.

\bibitem[Brown et~al.(2020)Brown, Mann, Ryder, Subbiah, Kaplan, Dhariwal,
  Neelakantan, Shyam, Sastry, Askell, et~al.]{brown2020language}
Brown, T., Mann, B., Ryder, N., Subbiah, M., Kaplan, J.~D., Dhariwal, P.,
  Neelakantan, A., Shyam, P., Sastry, G., Askell, A., et~al.
\newblock Language models are few-shot learners.
\newblock \emph{Advances in neural information processing systems},
  33:\penalty0 1877--1901, 2020.

\bibitem[Chase(2022)]{Chase_LangChain_2022}
Chase, H.
\newblock {LangChain}, 10 2022.
\newblock URL \url{https://github.com/hwchase17/langchain}.

\bibitem[Dalmia et~al.(2021)Dalmia, Yan, Raunak, Metze, and
  Watanabe]{dalmia2021searchable}
Dalmia, S., Yan, B., Raunak, V., Metze, F., and Watanabe, S.
\newblock Searchable hidden intermediates for end-to-end models of decomposable
  sequence tasks.
\newblock In \emph{Proceedings of the 2021 Conference of the North American
  Chapter of the Association for Computational Linguistics: Human Language
  Technologies}, pp.\  1882--1896, 2021.

\bibitem[D{\'e}fossez et~al.(2022)D{\'e}fossez, Copet, Synnaeve, and
  Adi]{defossez2022high}
D{\'e}fossez, A., Copet, J., Synnaeve, G., and Adi, Y.
\newblock High fidelity neural audio compression.
\newblock \emph{arXiv preprint arXiv:2210.13438}, 2022.

\bibitem[Devlin et~al.(2018)Devlin, Chang, Lee, and Toutanova]{devlin2018bert}
Devlin, J., Chang, M.-W., Lee, K., and Toutanova, K.
\newblock Bert: Pre-training of deep bidirectional transformers for language
  understanding.
\newblock \emph{arXiv preprint arXiv:1810.04805}, 2018.

\bibitem[Grabocka et~al.(2018)Grabocka, Schmidt~Thieme, and
  etc]{grabocka2018neuralwarp}
Grabocka, J., Schmidt~Thieme, L., and etc.
\newblock Neuralwarp: Time-series similarity with warping networks.
\newblock \emph{arXiv preprint arXiv:1812.08306}, 2018.

\bibitem[Gulcehre et~al.(2017)Gulcehre, Firat, Xu, Cho, and
  Bengio]{gulcehre2017integrating}
Gulcehre, C., Firat, O., Xu, K., Cho, K., and Bengio, Y.
\newblock On integrating a language model into neural machine translation.
\newblock \emph{Computer Speech \& Language}, 45:\penalty0 137--148, 2017.

\bibitem[Hayashi \& Watanabe(2020)Hayashi and Watanabe]{hayashi2020discretalk}
Hayashi, T. and Watanabe, S.
\newblock Discretalk: Text-to-speech as a machine translation problem.
\newblock \emph{arXiv preprint arXiv:2005.05525}, 2020.

\bibitem[Hosseini-Asl et~al.(2020)Hosseini-Asl, McCann, Wu, Yavuz, and
  Socher]{hosseini2020simple}
Hosseini-Asl, E., McCann, B., Wu, C.-S., Yavuz, S., and Socher, R.
\newblock A simple language model for task-oriented dialogue.
\newblock \emph{Advances in Neural Information Processing Systems},
  33:\penalty0 20179--20191, 2020.

\bibitem[Hsu et~al.(2021)Hsu, Bolte, Tsai, Lakhotia, Salakhutdinov, and
  Mohamed]{hsu2021hubert}
Hsu, W.-N., Bolte, B., Tsai, Y.-H.~H., Lakhotia, K., Salakhutdinov, R., and
  Mohamed, A.
\newblock Hubert: Self-supervised speech representation learning by masked
  prediction of hidden units.
\newblock \emph{IEEE/ACM Transactions on Audio, Speech, and Language
  Processing}, 29:\penalty0 3451--3460, 2021.

\bibitem[Huang et~al.(2021)Huang, Chen, Ren, Liu, Cui, and
  Zhao]{huang2021multi}
Huang, R., Chen, F., Ren, Y., Liu, J., Cui, C., and Zhao, Z.
\newblock Multi-singer: Fast multi-singer singing voice vocoder with a
  large-scale corpus.
\newblock In \emph{Proceedings of the 29th ACM International Conference on
  Multimedia}, pp.\  3945--3954, 2021.

\bibitem[Huang et~al.(2022{\natexlab{a}})Huang, Lam, Wang, Su, Yu, Ren, and
  Zhao]{huang2022fastdiff}
Huang, R., Lam, M.~W., Wang, J., Su, D., Yu, D., Ren, Y., and Zhao, Z.
\newblock Fastdiff: A fast conditional diffusion model for high-quality speech
  synthesis.
\newblock \emph{arXiv preprint arXiv:2204.09934}, 2022{\natexlab{a}}.

\bibitem[Huang et~al.(2022{\natexlab{b}})Huang, Ren, Liu, Cui, and
  Zhao]{huang2022generspeech}
Huang, R., Ren, Y., Liu, J., Cui, C., and Zhao, Z.
\newblock Generspeech: Towards style transfer for generalizable out-of-domain
  text-to-speech synthesis.
\newblock \emph{arXiv preprint arXiv:2205.07211}, 2022{\natexlab{b}}.

\bibitem[Huang et~al.(2022{\natexlab{c}})Huang, Zhao, Liu, Liu, Ren, Zhang, and
  He]{huang2022transpeech}
Huang, R., Zhao, Z., Liu, J., Liu, H., Ren, Y., Zhang, L., and He, J.
\newblock Transpeech: Speech-to-speech translation with bilateral perturbation.
\newblock \emph{arXiv preprint arXiv:2205.12523}, 2022{\natexlab{c}}.

\bibitem[Huang et~al.(2023{\natexlab{a}})Huang, Huang, Yang, Ren, Liu, Li, Ye,
  Liu, Yin, and Zhao]{huang2023make}
Huang, R., Huang, J., Yang, D., Ren, Y., Liu, L., Li, M., Ye, Z., Liu, J., Yin,
  X., and Zhao, Z.
\newblock Make-an-audio: Text-to-audio generation with prompt-enhanced
  diffusion models.
\newblock \emph{arXiv preprint arXiv:2301.12661}, 2023{\natexlab{a}}.

\bibitem[Huang et~al.(2023{\natexlab{b}})Huang, Dong, Wang, Hao, Singhal, Ma,
  Lv, Cui, Mohammed, Liu, et~al.]{huang2023language}
Huang, S., Dong, L., Wang, W., Hao, Y., Singhal, S., Ma, S., Lv, T., Cui, L.,
  Mohammed, O.~K., Liu, Q., et~al.
\newblock Language is not all you need: Aligning perception with language
  models.
\newblock \emph{arXiv preprint arXiv:2302.14045}, 2023{\natexlab{b}}.

\bibitem[Kaplan et~al.(2020)Kaplan, McCandlish, Henighan, Brown, Chess, Child,
  Gray, Radford, Wu, and Amodei]{kaplan2020scaling}
Kaplan, J., McCandlish, S., Henighan, T., Brown, T.~B., Chess, B., Child, R.,
  Gray, S., Radford, A., Wu, J., and Amodei, D.
\newblock Scaling laws for neural language models.
\newblock \emph{arXiv preprint arXiv:2001.08361}, 2020.

\bibitem[Kharitonov et~al.(2022)Kharitonov, Copet, Lakhotia, Nguyen, Tomasello,
  Lee, Elkahky, Hsu, Mohamed, Dupoux, et~al.]{kharitonov2022textless}
Kharitonov, E., Copet, J., Lakhotia, K., Nguyen, T.~A., Tomasello, P., Lee, A.,
  Elkahky, A., Hsu, W.-N., Mohamed, A., Dupoux, E., et~al.
\newblock textless-lib: A library for textless spoken language processing.
\newblock \emph{arXiv preprint arXiv:2202.07359}, 2022.

\bibitem[Liu et~al.(2020)Liu, Li, Zhao, and Jin]{liu2020multi}
Liu, F., Li, G., Zhao, Y., and Jin, Z.
\newblock Multi-task learning based pre-trained language model for code
  completion.
\newblock In \emph{Proceedings of the 35th IEEE/ACM International Conference on
  Automated Software Engineering}, pp.\  473--485, 2020.

\bibitem[Liu et~al.(2022{\natexlab{a}})Liu, Li, Ren, Chen, and
  Zhao]{liu2022diffsinger}
Liu, J., Li, C., Ren, Y., Chen, F., and Zhao, Z.
\newblock Diffsinger: Singing voice synthesis via shallow diffusion mechanism.
\newblock In \emph{Proceedings of the AAAI Conference on Artificial
  Intelligence}, 2022{\natexlab{a}}.

\bibitem[Liu et~al.(2022{\natexlab{b}})Liu, Liu, Kong, Mei, Zhao, Huang,
  Plumbley, and Wang]{liu2022separate}
Liu, X., Liu, H., Kong, Q., Mei, X., Zhao, J., Huang, Q., Plumbley, M.~D., and
  Wang, W.
\newblock Separate what you describe: language-queried audio source separation.
\newblock \emph{arXiv preprint arXiv:2203.15147}, 2022{\natexlab{b}}.

\bibitem[Luo \& Mesgarani(2019)Luo and Mesgarani]{luo2019conv}
Luo, Y. and Mesgarani, N.
\newblock Conv-tasnet: Surpassing ideal time--frequency magnitude masking for
  speech separation.
\newblock \emph{IEEE/ACM transactions on audio, speech, and language
  processing}, 27\penalty0 (8):\penalty0 1256--1266, 2019.

\bibitem[Nguyen et~al.(2023)Nguyen, Kharitonov, Copet, Adi, Hsu, Elkahky,
  Tomasello, Algayres, Sagot, Mohamed, et~al.]{nguyen2023generative}
Nguyen, T.~A., Kharitonov, E., Copet, J., Adi, Y., Hsu, W.-N., Elkahky, A.,
  Tomasello, P., Algayres, R., Sagot, B., Mohamed, A., et~al.
\newblock Generative spoken dialogue language modeling.
\newblock \emph{Transactions of the Association for Computational Linguistics},
  11:\penalty0 250--266, 2023.

\bibitem[Ouyang et~al.(2022)Ouyang, Wu, Jiang, Almeida, Wainwright, Mishkin,
  Zhang, Agarwal, Slama, Ray, et~al.]{ouyang2022training}
Ouyang, L., Wu, J., Jiang, X., Almeida, D., Wainwright, C., Mishkin, P., Zhang,
  C., Agarwal, S., Slama, K., Ray, A., et~al.
\newblock Training language models to follow instructions with human feedback.
\newblock \emph{Advances in Neural Information Processing Systems},
  35:\penalty0 27730--27744, 2022.

\bibitem[Radford et~al.(2022)Radford, Kim, Xu, Brockman, McLeavey, and
  Sutskever]{radford2022robust}
Radford, A., Kim, J.~W., Xu, T., Brockman, G., McLeavey, C., and Sutskever, I.
\newblock Robust speech recognition via large-scale weak supervision.
\newblock \emph{arXiv preprint arXiv:2212.04356}, 2022.

\bibitem[Raffel et~al.(2020)Raffel, Shazeer, Roberts, Lee, Narang, Matena,
  Zhou, Li, Liu, et~al.]{raffel2020exploring}
Raffel, C., Shazeer, N., Roberts, A., Lee, K., Narang, S., Matena, M., Zhou,
  Y., Li, W., Liu, P.~J., et~al.
\newblock Exploring the limits of transfer learning with a unified text-to-text
  transformer.
\newblock \emph{J. Mach. Learn. Res.}, 21\penalty0 (140):\penalty0 1--67, 2020.

\bibitem[Ren et~al.(2020)Ren, Hu, Tan, Qin, Zhao, Zhao, and
  Liu]{ren2020fastspeech}
Ren, Y., Hu, C., Tan, X., Qin, T., Zhao, S., Zhao, Z., and Liu, T.-Y.
\newblock Fastspeech 2: Fast and high-quality end-to-end text to speech.
\newblock \emph{arXiv preprint arXiv:2006.04558}, 2020.

\bibitem[Shen et~al.(2023)Shen, Song, Tan, Li, Lu, and
  Zhuang]{shen2023hugginggpt}
Shen, Y., Song, K., Tan, X., Li, D., Lu, W., and Zhuang, Y.
\newblock Hugginggpt: Solving ai tasks with chatgpt and its friends in
  huggingface.
\newblock \emph{arXiv preprint arXiv:2303.17580}, 2023.

\bibitem[Svyatkovskiy et~al.(2019)Svyatkovskiy, Zhao, Fu, and
  Sundaresan]{svyatkovskiy2019pythia}
Svyatkovskiy, A., Zhao, Y., Fu, S., and Sundaresan, N.
\newblock Pythia: Ai-assisted code completion system.
\newblock In \emph{Proceedings of the 25th ACM SIGKDD international conference
  on knowledge discovery \& data mining}, pp.\  2727--2735, 2019.

\bibitem[Thoppilan et~al.(2022)Thoppilan, De~Freitas, Hall, Shazeer,
  Kulshreshtha, Cheng, Jin, Bos, Baker, Du, et~al.]{thoppilan2022lamda}
Thoppilan, R., De~Freitas, D., Hall, J., Shazeer, N., Kulshreshtha, A., Cheng,
  H.-T., Jin, A., Bos, T., Baker, L., Du, Y., et~al.
\newblock Lamda: Language models for dialog applications.
\newblock \emph{arXiv preprint arXiv:2201.08239}, 2022.

\bibitem[Touvron et~al.(2023)Touvron, Lavril, Izacard, Martinet, Lachaux,
  Lacroix, Rozi{\`e}re, Goyal, Hambro, Azhar, et~al.]{touvron2023llama}
Touvron, H., Lavril, T., Izacard, G., Martinet, X., Lachaux, M.-A., Lacroix,
  T., Rozi{\`e}re, B., Goyal, N., Hambro, E., Azhar, F., et~al.
\newblock Llama: Open and efficient foundation language models.
\newblock \emph{arXiv preprint arXiv:2302.13971}, 2023.

\bibitem[Wang et~al.(2022)Wang, Cornell, Choi, Lee, Kim, and
  Watanabe]{wang2022tf}
Wang, Z.-Q., Cornell, S., Choi, S., Lee, Y., Kim, B.-Y., and Watanabe, S.
\newblock Tf-gridnet: Making time-frequency domain models great again for
  monaural speaker separation.
\newblock \emph{arXiv preprint arXiv:2209.03952}, 2022.

\bibitem[Wei et~al.(2021)Wei, Bosma, Zhao, Guu, Yu, Lester, Du, Dai, and
  Le]{wei2021finetuned}
Wei, J., Bosma, M., Zhao, V.~Y., Guu, K., Yu, A.~W., Lester, B., Du, N., Dai,
  A.~M., and Le, Q.~V.
\newblock Finetuned language models are zero-shot learners.
\newblock \emph{arXiv preprint arXiv:2109.01652}, 2021.

\bibitem[Wu et~al.(2023)Wu, Yin, Qi, Wang, Tang, and Duan]{wu2023visual}
Wu, C., Yin, S., Qi, W., Wang, X., Tang, Z., and Duan, N.
\newblock Visual chatgpt: Talking, drawing and editing with visual foundation
  models.
\newblock \emph{arXiv preprint arXiv:2303.04671}, 2023.

\bibitem[Wu et~al.(2021)Wu, Jia, Wang, Dou, Duan, and Deng]{wu2021imitating}
Wu, H., Jia, J., Wang, H., Dou, Y., Duan, C., and Deng, Q.
\newblock Imitating arbitrary talking style for realistic audio-driven talking
  face synthesis.
\newblock In \emph{Proceedings of the 29th ACM International Conference on
  Multimedia}, pp.\  1478--1486, 2021.

\bibitem[Xin et~al.(2022)Xin, Yang, and Zou]{xin2022audio}
Xin, Y., Yang, D., and Zou, Y.
\newblock Audio pyramid transformer with domain adaption for weakly supervised
  sound event detection and audio classification.
\newblock \emph{Proc. Interspeech 2022}, pp.\  1546--1550, 2022.

\bibitem[Yang et~al.(2022)Yang, Yu, Wang, Wang, Weng, Zou, and
  Yu]{yang2022diffsound}
Yang, D., Yu, J., Wang, H., Wang, W., Weng, C., Zou, Y., and Yu, D.
\newblock Diffsound: Discrete diffusion model for text-to-sound generation.
\newblock \emph{arXiv preprint arXiv:2207.09983}, 2022.

\bibitem[Ye et~al.(2021)Ye, Wang, Yang, and Zou]{ye2021improving}
Ye, Z., Wang, H., Yang, D., and Zou, Y.
\newblock Improving the performance of automated audio captioning via
  integrating the acoustic and semantic information.
\newblock \emph{arXiv preprint arXiv:2110.06100}, 2021.

\bibitem[Ye et~al.(2023)Ye, Jiang, Ren, Liu, He, and Zhao]{ye2023geneface}
Ye, Z., Jiang, Z., Ren, Y., Liu, J., He, J., and Zhao, Z.
\newblock Geneface: Generalized and high-fidelity audio-driven 3d talking face
  synthesis.
\newblock \emph{arXiv preprint arXiv:2301.13430}, 2023.

\bibitem[Zeghidour et~al.(2021)Zeghidour, Luebs, Omran, Skoglund, and
  Tagliasacchi]{zeghidour2021soundstream}
Zeghidour, N., Luebs, A., Omran, A., Skoglund, J., and Tagliasacchi, M.
\newblock Soundstream: An end-to-end neural audio codec.
\newblock \emph{IEEE/ACM Transactions on Audio, Speech, and Language
  Processing}, 30:\penalty0 495--507, 2021.

\bibitem[Zhang et~al.(2022{\natexlab{a}})Zhang, Roller, Goyal, Artetxe, Chen,
  Chen, Dewan, Diab, Li, Lin, et~al.]{zhang2022opt}
Zhang, S., Roller, S., Goyal, N., Artetxe, M., Chen, M., Chen, S., Dewan, C.,
  Diab, M., Li, X., Lin, X.~V., et~al.
\newblock Opt: Open pre-trained transformer language models.
\newblock \emph{arXiv preprint arXiv:2205.01068}, 2022{\natexlab{a}}.

\bibitem[Zhang et~al.(2022{\natexlab{b}})Zhang, Cong, Xue, Xie, Zhu, and
  Bi]{zhang2022visinger}
Zhang, Y., Cong, J., Xue, H., Xie, L., Zhu, P., and Bi, M.
\newblock Visinger: Variational inference with adversarial learning for
  end-to-end singing voice synthesis.
\newblock In \emph{ICASSP 2022-2022 IEEE International Conference on Acoustics,
  Speech and Signal Processing (ICASSP)}, pp.\  7237--7241. IEEE,
  2022{\natexlab{b}}.

\end{thebibliography}
